\documentclass[conference]{IEEEtran}
\IEEEoverridecommandlockouts
\usepackage{adjustbox} 
\usepackage{cite}
\usepackage{amsmath,amssymb,amsfonts}
\usepackage{algorithmic}
\usepackage{graphicx}
\usepackage{textcomp}
\usepackage{xcolor}
\usepackage{url}
\usepackage{graphicx}
\usepackage{multirow}
\usepackage{graphicx}
\usepackage{caption}
\usepackage{booktabs}
\usepackage{xcolor}
\usepackage{amsmath}
\usepackage{array}
\usepackage{hyperref}

\def\BibTeX{{\rm B\kern-.05em{\sc i\kern-.025em b}\kern-.08em
    T\kern-.1667em\lower.7ex\hbox{E}\kern-.125emX}}
\begin{document}

\title{BRAINS: A Retrieval-Augmented System \\for Alzheimer’s Detection and Monitoring}

\author{%
\textbf{Rajan Das Gupta}$^{1,4}$  \quad
\textbf{Md Kishor Morol}$^{4,\dag}$\thanks{† Corresponding author} \quad
\textbf{Nafiz Fahad}$^{2,4}$ \quad \\
 \textbf{Md Tanzib Hosain}$^{1,4}$ \quad
    \textbf{Sumaya Binte Zilani Choya}$^{4,4}$ \quad
    \textbf{Md Jakir Hossen}$^{2,4,\dag}$
 \\
 [1ex]
 $^{1}$American International University-Bangladesh \quad 
  $^{2}$Multimedia University \quad\\
 $^{3}$George Mason University \quad
 $^{4}$ELITE Research Lab \quad
 \\
 \texttt{18-36304-1@student.aiub.edu, kishormorol@ieee.org}\\ \texttt{nafiz.fahad@student.mmu.edu.my, 20-42737-1@student.aiub.edu}\\ \texttt{schoya@gmu.edu, jakir.hossen@mmu.edu.my}
}

\maketitle

\begin{abstract}
 As the global burden of Alzheimer's disease (AD) continues to grow, early and accurate detection has become increasingly critical, especially in regions with limited access to advanced diagnostic tools. We propose BRAINS (Biomedical Retrieval-Augmented Intelligence for Neurodegeneration Screening) to address this challenge. This novel system harnesses the powerful reasoning capabilities of Large Language Models (LLMs) for Alzheimer's detection and monitoring. BRAINS features a dual-module architecture: a cognitive diagnostic module and a case-retrieval module. The Diagnostic Module utilises LLMs fine-tuned on cognitive and neuroimaging datasets—including MMSE, CDR scores, and brain volume metrics—to perform structured assessments of Alzheimer's risk. Meanwhile, the Case Retrieval Module encodes patient profiles into latent representations and retrieves similar cases from a curated knowledge base. These auxiliary cases are fused with the input profile via a Case Fusion Layer to enhance contextual understanding. The combined representation is then processed with clinical prompts for inference. Evaluations on real-world datasets demonstrate BRAINS effectiveness in classifying disease severity and identifying early signs of cognitive decline. This system not only shows strong potential as an assistive tool for scalable, explainable, and early-stage Alzheimer's disease detection, but also offers hope for future applications in the field.
\end{abstract}

\begin{IEEEkeywords}
Alzheimer’s Disease, Retrieval Augmented Generation (RAG), Cognitive Decline Detection, Clinical Decision Support\end{IEEEkeywords}

\section{Introduction}

Alzheimer's disease (AD), the most common form of dementia, represents a complex and progressive neurodegenerative disorder that significantly impairs memory, cognition, and behaviour \cite{imbimbo2021accelerating}. According to the World Health Organization (WHO), over 55 million people worldwide are currently living with dementia, with Alzheimer's disease accounting for up to 70\% of these cases. Alarmingly, this number is projected to rise to 139 million by 2050, driven primarily by ageing populations and delayed diagnoses \cite{miller2018astrocyte}. As noted by the National Institute on Aging (NIA), Alzheimer's is currently among the leading causes of disability and dependency among older adults. However, it remains vastly underdiagnosed, particularly in low-resource settings where access to specialized neurological assessment is limited \cite{petersen2010adni}.
\vspace{1mm}
\par Despite decades of research, early and reliable detection of Alzheimer's disease remains a critical challenge. Traditional diagnostic methods, including neuropsychological testing, MRI-based brain volume analysis~\cite{marcus2007open}, and clinical rating scales such as the Mini-Mental State Examination (MMSE) and Clinical Dementia Rating (CDR)~\cite{morris1993cdr}, are resource-intensive and require domain expertise \cite{jack2018nia}. Moreover, access to neuroimaging and specialized interpretation remains scarce in economically disadvantaged regions, leading to disparities in early diagnosis and care delivery~\cite{weiner2015update,folstein1975mmse}.
\vspace{1mm}
\par Diagnosing and managing Alzheimer's disease is further complicated by the brain's structural complexity and the heterogeneity of cognitive decline patterns. Small morphological changes, such as cortical thinning or hippocampal atrophy, are often difficult to quantify and interpret, even for experienced clinicians~\cite{reuben2021predicting}. Additionally, real-world assessment data, such as brain volume estimates (eTIV, nWBV), MMSE/CDR scores, and demographic information (Table~\ref{tab:alz_features}), are inherently variable and incomplete across populations~\cite{yang2023knowledge,singhal2023expert,luo2024clinicalt5}. These challenges underscore the need for intelligent systems integrating heterogeneous, multimodal data to aid early detection and disease monitoring~\cite{zeng2024longcontext,gao2023reta,zhang2024cogagent}.
\vspace{1mm}
\par Recent advances in artificial intelligence (AI) particularly large language models (LLMs) offer transformative potential for assisting clinicians in synthesizing such complex data. While previous AI systems in neurology have relied on rigid feature engineering or domain-specific heuristics, LLMs~\cite{chen2024pmcllama,chowdhery2023palm,achiam2023gpt,li2023medpalm,touvron2023llama} provide a generalizable, prompt-driven framework for reasoning over structured clinical input. However, most current systems lack case-based contextual reasoning, interpretability, and robustness to real-world data variability.
\vspace{1mm}

\begin{table*}[h]
\caption{Key Features for Alzheimer’s Disease Diagnosis}
\centering
\begin{tabular}{p{8.5cm}|p{8.5cm}}
\hline
\textbf{Feature} & \textbf{Clinical Relevance} \\
\hline
MMSE & Global cognitive function \\
\hline
CDR & Dementia severity rating \\
\hline
eTIV & Intracranial volume normalization \\
\hline
nWBV & Brain atrophy indicator \\
\hline
Hippocampal Volume & Memory-related region atrophy \\
\hline
Amygdala Volume & Emotion processing region \\
\hline
Ventricular Volume & Enlarged in AD progression \\
\hline
Temporal Thickness & Cortical atrophy biomarker \\
\hline
WMH & Marker for vascular pathology \\
\hline
Age & Primary risk factor \\
\hline
Education Level & Reflects cognitive reserve \\
\hline
Gender & Influences disease manifestation \\
\hline
APOE Genotype & Genetic predisposition ($\epsilon4$ allele) \\
\hline
MoCA Score & MCI screening alternative \\
\hline
GDS Score & Screens comorbid depression \\
\hline
\end{tabular}
\label{tab:alz_features}
\end{table*}

\par To address these limitations, we introduce \textbf{BRAINS} (\textit{Biomedical Retrieval-Augmented Intelligence for Neurodegeneration Screening})---a novel, retrieval-augmented diagnostic framework that leverages the reasoning capabilities of LLMs alongside case-based retrieval and neurocognitive data fusion. BRAINS adopts a dual-phase architecture: the Diagnostic Module is pre-trained with neurocognitive assessment records to encode foundational knowledge of Alzheimer's progression. In contrast, the Case Retrieval Module encodes input features to retrieve semantically relevant historical cases from a clinical knowledge base. These auxiliary cases are integrated through a Case Fusion Layer, enabling context-aware reasoning that enhances interpretability and prediction accuracy.
\vspace{1mm}
\par We construct and validate BRAINS on a real-world dataset comprising MMSE, CDR, brain volumetric measures (eTIV, nWBV), and demographic variables to evaluate the system's practical utility. Our experiments demonstrate that BRAINS achieves superior performance in detecting and staging Alzheimer's disease compared to baseline models, while also providing explainable outputs aligned with clinical insights. We argue that BRAINS not only contributes a robust technical advancement but also represents a scalable solution for dementia screening in both high-resource hospitals and underserved settings where early intervention is most critical.

\section{Methodology}
In this section, we introduce \textbf{BRAINS} (Biomedical Retrieval-Augmented Intelligence for Neurodegeneration Screening), a system designed to support Alzheimer’s disease diagnosis by leveraging clinical guidelines in neurocognitive assessment and brain imaging. We begin by describing our dataset, which includes demographic, cognitive, and neuroanatomical features such as MMSE, CDR, eTIV, and nWBV. We then present the architecture of BRAINS, which combines a Diagnostic Module for cognitive evaluation with a Case Retrieval Module that integrates relevant patient histories. These components are fused via a Case Fusion Layer, enabling context-aware, explainable predictions aligned with real-world clinical workflows.

\subsection{Data Preparing}\label{AA}
The objective of pre-training is to equip the foundational model with prior domain knowledge by exposing it to a broad corpus of relevant information. This stage involves training on a large volume of unlabeled text to help the model internalize domain-specific concepts before fine-tuning on structured diagnostic tasks. In our case, we aim for the BRAINS model to acquire baseline understanding of Alzheimer's-related terminology, cognitive scoring systems, and clinical reasoning prior to task-specific adaptation.
\vspace{1mm}
\par To support this, we curated a pre-training dataset consisting of textual reports and summaries derived from relevant Alzheimer’s disease report~\cite{frisoni2010clinical,weiner2015update,marcus2007open}, neurocognitive evaluations (e.g., MMSE, CDR)~\cite{morris1993clinical}, and structured annotations from clinical databases such as NACC and ADNI~\cite{nacc2023,adni2023}. Since the model is not designed to process images directly, we applied regularization by removing sentences that referenced visual elements. Specifically, we filtered out any sentences containing terms like “Figure” or “see image” to ensure consistency in textual input. This pre-training phase enhances the model’s ability to reason over structured brain health data in subsequent diagnostic tasks.

\subsection{Alzheimer's Disease Dataset}
To support the diagnosis and staging of Alzheimer's disease, we curated a further clinical dataset with 1105 patient records collected from medical institutions. Each subject is categorized as 1. Early-Onset Alzheimer’s Disease, 2. Late-Onset Alzheimer’s Disease, 3. Familial Alzheimer’s Disease, 4. Sporadic Alzheimer’s Disease, 5. Atypical Alzheimer’s Disease based on MMSE, CDR, and neuroimaging-derived metrics.

\par The dataset includes features: MMSE, CDR, eTIV, nWBV, age, gender, handedness, education, and socioeconomic status. All entries were pre-processed by normalization, encoding, and outlier removal to ensure quality and consistency for downstream modeling with BRAINS.
\begin{figure*}[h]
\centering
\includegraphics[width=\textwidth]{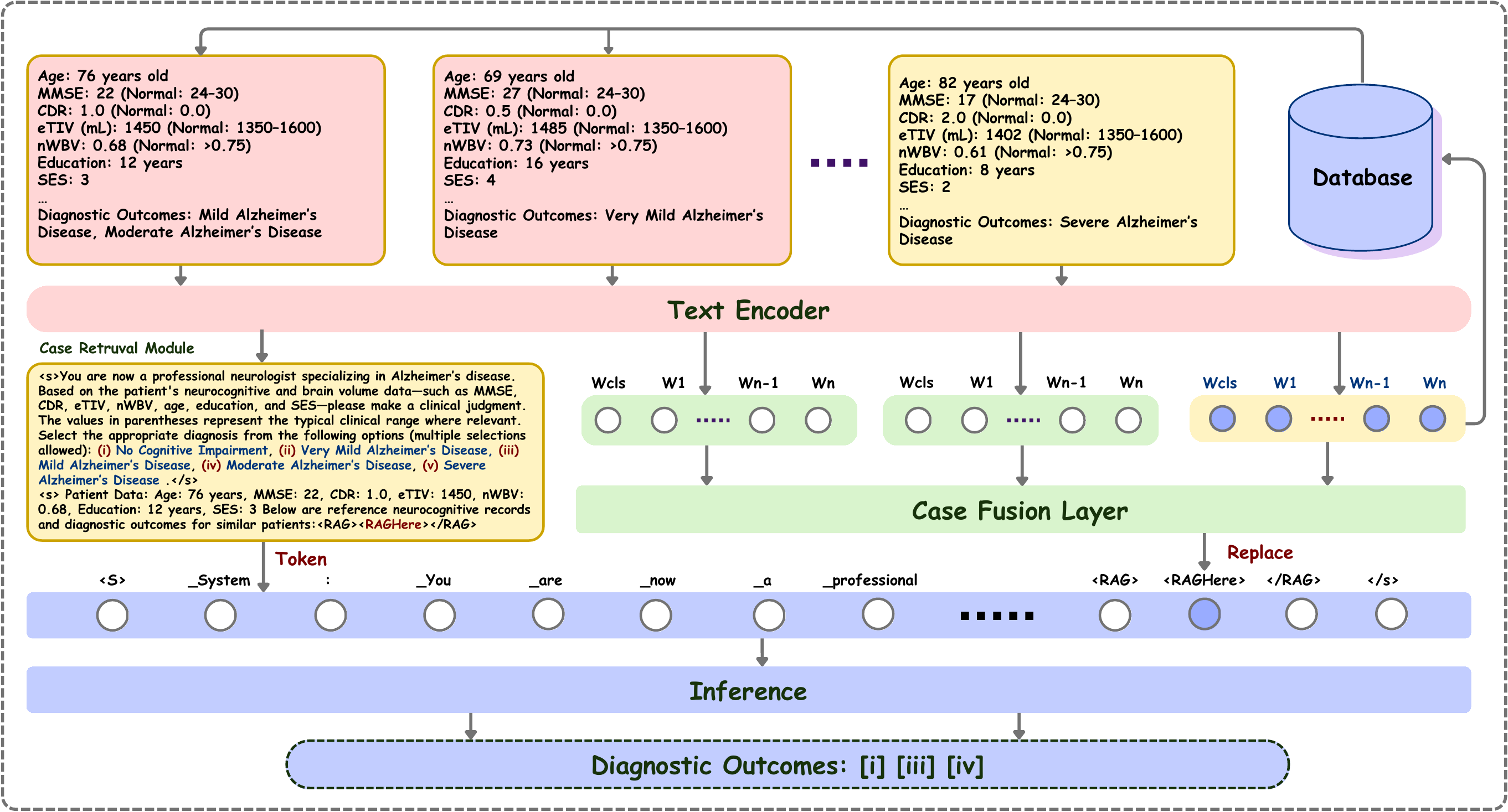} 
\caption{BRAINS architecture for Alzheimer’s diagnosis. The input case is encoded and used to retrieve similar neurocognitive records. Retrieved cases are fused with the input via the Case Fusion Layer, replacing the \texttt{<RAGHere>} token in the prompt. The fused representation is then passed to the LLM for inference and explanation.}
\label{fig:brains_architecture}
\end{figure*}

\subsection{Model Architecture}
As illustrated in Figure~\ref{fig:brains_architecture}, the proposed \textbf{BRAINS} framework consists of two core components: the \textit{Diagnostic Module} and the \textit{Case Retrieval Module}. Upon receiving a new patient case—comprising neurocognitive scores (e.g., MMSE, CDR), demographic data, and structural brain metrics (e.g., nWBV, eTIV)—the Case Retrieval Module first encodes the input and retrieves the top-KKK most clinically relevant auxiliary cases from a curated knowledge base. These retrieved cases are then paired with the input as (T, R)(T, R)(T, R), where TTT represents the target case and RRR denotes the reference cases. This case pair is subsequently passed to the Diagnostic Module, which performs reasoning over the joint representation to assess the likelihood and stage of Alzheimer's disease. By integrating contextual information from similar historical cases, BRAINS enhances interpretability and diagnostic robustness. The following sections detail each module's architecture and operational flow.

\subsection{Case Retrieval Module}

To enhance the precision and contextual awareness of Alzheimer's disease diagnosis, the proposed \textbf{BRAINS} system integrates retrieval-augmented inference by leveraging a structured memory of historical neurocognitive cases. Rather than relying solely on generalizable reasoning from pre-trained knowledge, BRAINS employs a Case Retrieval Module to dynamically fetch semantically similar patient profiles from a vectorized clinical case database. This augmentation facilitates nuanced decision-making, especially in the presence of heterogeneous or borderline cognitive features.
\vspace{1mm}
\par The Case Retrieval Module comprises two core components: a curated neurocognitive case database and a high-dimensional text encoder. Each entry in the training dataset—including features such as MMSE, CDR, age, eTIV, nWBV, and education—is first tokenized and encoded using a clinical-domain-adapted text encoder. The resulting hidden representation $\mathbf{w}_{\text{[CLS]}}$, which summarizes the input case, is stored in a FAISS-based vector database \cite{douze2025faisslibrary}.
\vspace{1mm}
\par During training and inference, the input case $T$ is encoded into its hidden representation and queried against the FAISS vector store using cosine similarity. The top $1K$ most similar historical cases are retrieved, followed by a reranking module that reorders them based on a fine-tuned scoring mechanism. The highest scoring $K$ auxiliary cases $R = \{r_0, r_1, \ldots, r_{K-1}\}$ are then selected to form a case pair $(T, R)$, where the target case $T$ is jointly reasoned over with the retrieved support cases $R$. This approach allows \textbf{BRAINS} to directly incorporate population-level cognitive trends and structural brain biomarkers into its diagnostic inference, ultimately improving robustness and explainability in clinical settings.

\begin{table*}[h]
\centering
\caption{Performance comparison of \texttt{LLaMA2-13B}, RAG variants, and the proposed BRAINS system across all, single, double, and triple case types.}
\renewcommand{\arraystretch}{1.2}
\setlength{\tabcolsep}{6pt}
\begin{tabular}{l|c|c|c|c|c|c|c|c|c|c|c}
\hline
\multirow{2}{*}{\textbf{Model}} 
& \multicolumn{2}{c|}{\textbf{All}} 
& \multicolumn{3}{c|}{\textbf{Single}} 
& \multicolumn{3}{c|}{\textbf{Double}} 
& \multicolumn{3}{c}{\textbf{Triple}} \\ \cline{2-12}
& Correct & F1 & Prec. & Recall & F1 & Prec. & Recall & F1 & Prec. & Recall & F1 \\ \hline

\textbf{LLaMA2-13B} \\\hline
Five-shot 
& 0.335 & 0.339 & 0.000 & 0.000 & 0.000
& 0.299 & 0.719 & 0.423 
& 0.421 & 0.980 & 0.591  \\

Fine-tuning 
& 0.600 & 0.538 & 0.657 & 0.728 & 0.692
& 0.468 & 0.474 & 0.471 
& 0.643 & 0.281 & 0.391 \\

w/o standard 
& 0.454 & 0.376 & 0.645 & 0.513 & 0.571
& 0.290 & 0.500 & 0.361 
& 0.250 & 0.063 & 0.100  \\ \hline

\textbf{RAG}\\\hline
RAG-1 
& 0.712 & 0.731 & 0.766 & 0.540 
& 0.619 & 0.703 & 0.824 
& 0.802 & 0.774 & 0.981 
& 0.863  \\

RAG-2 
& 0.727 & 0.755 & 0.790 & 0.572 
& 0.664 & 0.660 & 0.921  
& 0.769 & 0.727 & 0.975 & 0.842 
\\ \hline

\textbf{BRAINS} 
& \textbf{0.773} & \textbf{0.819} & \textbf{0.784} & \textbf{0.731} 
& \textbf{0.740} & \textbf{0.711} & \textbf{0.875} 
& \textbf{0.810} & \textbf{0.931} & \textbf{0.911} 
& \textbf{0.929} \\ \hline

\end{tabular}
\label{tab:performance}
\end{table*}

\subsection{Diagnostic Module}

Upon retrieval, the BRAINS framework forms a case pair $(T, R)$, where $T$ represents the input patient profile and $R = \{r_0, r_1, ..., r_{K-1}\}$ denotes the top-$K$ semantically relevant neurocognitive cases retrieved from the clinical memory. The Diagnostic Module consists of two key components: the \textit{Case Fusion Layer} and the \textit{Inference LLM}.
\vspace{1mm}
\par To overcome the context-length limitations of large language models, the Case Fusion Layer aggregates representations of the retrieved cases. Each auxiliary case $r_i$ is encoded into hidden vectors $w_i = \{w_{\text{[CLS]}_i}, w_{1i}, \ldots, w_{ni}\}$, and the input case is encoded as $t = \{t_{\text{[CLS]}}, t_1, \ldots, t_n\}$. A cross-attention mechanism, based on the standard Transformer formulation~\cite{vaswani2017attention}, is employed to align and integrate contextual information across the retrieved examples:

\begin{equation}
\text{Attn}(Q, K, V) = \text{softmax}\left( \frac{QK^\top}{\sqrt{d_k}} \right)V
\end{equation}

\noindent where $Q = W_Q t_{\text{[CLS]}}$, $K = W_K A$, $V = W_K A$, and $A$ denotes the concatenated matrix of retrieved case vectors:

\[
A = [w_{\text{[CLS]}_0}, w_{10}, \ldots, w_{n0}, w_{\text{[CLS]}_1}, \ldots, w_{n(K-1)}]
\]

The resulting case fusion vector replaces the special \texttt{<RAGHere>} token within the prompt embedding sequence $P = \{p_{\texttt{<s>}}, p_{\texttt{System}}, \ldots, p_{\texttt{<RAGHere>}}, \ldots, p_{\texttt{</s>}}\}$. The fused sequence is then passed to the Inference LLM for prediction.

During pre-training, we adopt a next-token prediction strategy to enable the model to acquire foundational neurocognitive knowledge. Fine-tuning is conducted with supervised objectives, where the loss is computed solely on the assistant's response, refining the model’s focus and performance in Alzheimer’s disease prediction tasks.

\section{Experiments}

In this section, we present the experimental framework employed to evaluate the efficacy of our proposed system. This framework is grounded in clinically relevant neurocognitive datasets, with a focus on early-stage Alzheimer's disease detection and multi-faceted cognitive impairment classification. The experimental design is meticulously structured to probe retrieval-augmented reasoning performance and generalization across single-clue, dual-clue, and complex multi-clue diagnostic cases. Implementation details, data preprocessing strategies, and evaluation protocols are aligned with established neuroscience benchmarks~\cite{guo2024heart} to ensure reproducibility and relevance. The results provide quantitative insight into the system's diagnostic accuracy and interpretability under variable cognitive load conditions.

\subsection{Pre-training Setting}\label{ITH}
We adopt the \texttt{LLaMA2-13B}~\cite{touvron2023llama} model as our foundational large language model, further fine-tuned on clinically curated neurocognitive corpora involving Alzheimer's disease, mild cognitive impairment (MCI), and related neurodegenerative conditions. Training is conducted over 10 epochs with a batch size of 64, using the AdamW optimizer~\cite{loshchilov2017decoupled} and a learning rate set to $1 \times 10^{-4}$. To ensure stable convergence, we apply 1{,}000 warm-up steps. The token block size is fixed at 2048 to accommodate high-resolution case narratives that integrate structured cognitive assessments (e.g., MMSE, CDR), neuroimaging-derived biomarkers (e.g., hippocampal atrophy, cortical thinning), and demographic variables (e.g., age, gender, education). This fine-tuning strategy equips the model with domain-specific reasoning capabilities necessary for downstream neurodiagnostic inference.

\subsection{Fine-tuning Setting}\label{FAT}
Building upon prior foundational training, we repurpose the pretrained neurodiagnostic language model as the backbone for inference. To ensure alignment between retrieval and reasoning, the same model architecture is deployed as the text encoder within the Case Retrieval Module, thereby preserving representational consistency. For reranking retrieved support cases, we integrate the \texttt{bge-reranker-large}~\cite{xiao2023bgereranker} to refine case relevance via dense semantic scoring. At inference time, we retrieve the top-$K=5$ neurocognitive profiles from the vector database, grounded in prior subjects exhibiting comparable biomarker and cognitive test signatures.
\vspace{1mm}
\par To promote positional robustness and mitigate overfitting to fixed retrieval slots, we introduce a dynamic masking strategy during training, where $m \in [0, 4]$ auxiliary cases are randomly masked in each training iteration. The model is trained for 15 epochs with a batch size of 4 using the AdamW optimizer~\cite{loshchilov2017decoupled} and a learning rate of $1 \times 10^{-5}$. To improve training efficiency and parameter efficiency, we adopt Low-Rank Adaptation (LoRA)~\cite{hu2022lora}, configured with $\alpha=32$ and $r=8$, enabling scalable fine-tuning within computational constraints while retaining high domain-specific fidelity.

\section{Results}\label{FAT}
As presented in Table~II, the efficacy of the proposed
\textsc{BRAINS} model is validated through rigorous empirical
evaluation within the domain of neurocognitive disorder inference.
In the baseline setting using \texttt{LLaMA2-13B}, the Five-shot prompting
paradigm can generate plausible outputs, yet it fails to yield
reliable diagnostic accuracy in complex multi-label neurocognitive
cases.

Upon fine-tuning the model on structured clinical
texts---enriched with cognitive metrics such as MMSE, CDR,
and volumetric MRI-derived features---a substantial improvement
of \textbf{26.50\%} in accuracy is observed. The absence of
these standardized neuro-biomarkers during input leads to a
pronounced degradation in model performance, especially in
complex classification settings involving co-occurring conditions.
Further gains are achieved by incorporating retrieval-augmented
generation (RAG), where one or more relevant cases are
retrieved to assist inference. With a single retrieved case
(RAG-1), accuracy increases from \textbf{60.00\%} to
\textbf{71.20\%}, and additional retrieved profiles provide
incremental benefits. However, retrieving more than two cases
leads to token length constraints that inhibit reasoning due
to limited context window sizes. This bottleneck is effectively
overcome by the \textsc{BRAINS} model’s case fusion mechanism,
which supports the integration of up to five auxiliary cases,
achieving an accuracy of \textbf{77.30\%}.

Under the Five-shot configuration, the model achieves
high recall (\textbf{98.00\%}) in multi-pathology scenarios
but suffers from low F1 (\textbf{59.10\%}) due to poor precision.
In contrast, single-label prediction tasks reveal a complete
breakdown (\textbf{F1 = 0.00\%}), suggesting an overly
conservative decision boundary. This highlights the limitations
of zero- or Five-shot generalization for complex neurological
inference. Fine-tuning improves this imbalance, while the
\textsc{BRAINS} model surpasses both token and reasoning
limitations, delivering robust and interpretable predictions
across varying levels of diagnostic complexity.

\section{Conclusion}

This study introduces BRAINS, a novel foundation model architected specifically for early-stage alzheimer's disorder screening. Tailored for the analysis of core neurological report data—such as cognitive assessments (e.g., MMSE, CDR), speech and behaviour logs, and structural brain imaging summaries—BRAINS is designed to support clinical reasoning, particularly for less experienced practitioners in neurology and geriatric medicine. By integrating retrieval-augmented generation (RAG) mechanisms, BRAINS substantially improves diagnostic precision for multi-morbidity inference tasks. In our benchmark evaluation on mild cognitive impairment and Alzheimer-type dementia classification, BRAINS achieves an accuracy of \textbf{77.30\%,} significantly outperforming the baseline large language model, which attains only \textbf{45.40\%}. We posit that BRAINS represents a foundational shift in scalable, interpretable, and data-efficient brain health modelling, with the potential to generalise across a broad spectrum of neurological diagnostic domains.

\bibliographystyle{ieeetr}
\bibliography{ref}

\end{document}